\documentclass{article}

\usepackage[preprint]{neurips_2023}

\usepackage[utf8]{inputenc} 
\usepackage[T1]{fontenc}    
\usepackage{hyperref}       
\usepackage{url}            
\usepackage{booktabs}       
\usepackage{amsfonts}       
\usepackage{nicefrac}       
\usepackage{microtype}      
\usepackage{xcolor}         
\usepackage{colortbl}

\usepackage{subcaption}
\usepackage{graphicx}
\usepackage{amsmath}
\usepackage{verbatim}
\usepackage{algorithm}
\usepackage{algpseudocode}
\usepackage{amsmath}
\usepackage{amssymb}
\usepackage{multirow}
\usepackage{makecell}
\usepackage{booktabs}
\usepackage{stfloats}
\usepackage[misc]{ifsym}
\usepackage{arydshln}
\usepackage{CJKutf8}
\usepackage{color}
\usepackage{bbding}
\usepackage{wrapfig}
\definecolor{metric}{RGB}{137,207,241}
\definecolor{dialogue}{RGB}{214,223,229}
\definecolor{charglm}{RGB}{0,127,219}
\definecolor{minimax}{RGB}{255,179,16}

\newcommand*\samethanks[1][\value{footnote}]{\footnotemark[#1]}
\newcommand{\logo}{\includegraphics[height=1em,trim=0 3em 0 0]{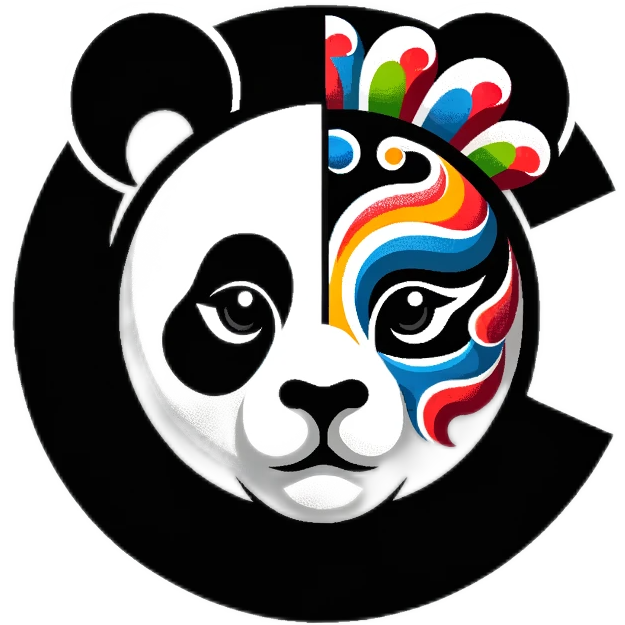}}

\title{\logo haracterGLM: Customizing Chinese Conversational\\AI Characters with Large Language Models}

\author{
    Jinfeng Zhou\textsuperscript{\rm 1}\thanks{Equal contribution.}\quad
    Zhuang Chen\textsuperscript{\rm 1}\samethanks{}\quad
    Dazhen Wan\textsuperscript{\rm 2}\samethanks{}\quad
    Bosi Wen\textsuperscript{\rm 1}\samethanks{}\quad
    Yi Song\textsuperscript{\rm 1}\samethanks{}\quad
    Jifan Yu\textsuperscript{\rm 3}\\
    \textbf{Yongkang Huang\textsuperscript{\rm 2}\quad
    Libiao Peng\textsuperscript{\rm 2}\quad
    Jiaming Yang\textsuperscript{\rm 2}\quad
    Xiyao Xiao\textsuperscript{\rm 2}\quad
    Sahand Sabour\textsuperscript{\rm 1}} \\
    \textbf{Xiaohan Zhang\textsuperscript{\rm 4}\quad
    Wenjing Hou\textsuperscript{\rm 5}\quad
    Yijia Zhang\textsuperscript{\rm 2}\quad
    Yuxiao Dong\textsuperscript{\rm 4,6}} \\ 
    \textbf{Jie Tang\textsuperscript{\rm 4,6}\quad
    Minlie Huang\textsuperscript{\rm 1,2}\thanks{Corresponding author.}}\\
    \textsuperscript{\rm 1}The CoAI Group, DCST, Tsinghua University \quad \textsuperscript{\rm 2}Lingxin AI \\
    \textsuperscript{\rm 3}Dept. of Computer SCi. \& Tech., Tsinghua University \quad
    \textsuperscript{\rm 4}Zhipu AI \\
    \textsuperscript{\rm 5}Renmin University of China \quad
    \textsuperscript{\rm 6}Knowledge Engineering Group, DCST, Tsinghua University\\
    \texttt{zjf23@mails.tsinghua.edu.cn, zhchen-nlp@mail.tsinghua.edu.cn}
}

\begin{document}

\maketitle

\begin{abstract}

In this paper, we present CharacterGLM, a series of models built upon ChatGLM, with model sizes ranging from 6B to 66B parameters. Our CharacterGLM is designed for generating Character-based Dialogues (CharacterDial), which aims to equip a conversational AI system with character customization for satisfying people’s inherent social desires and emotional needs. On top of CharacterGLM, we can customize various AI characters or social agents by configuring their attributes (identities, interests, viewpoints, experiences, achievements, social relationships, etc.) and behaviors (linguistic features, emotional expressions, interaction patterns, etc.). Our model outperforms most mainstream close-source large langauge models, including the GPT series, especially in terms of consistency, human-likeness, and engagement according to manual evaluations. We will release our 6B version of CharacterGLM and a subset of training data to facilitate further research development in the direction of character-based dialogue generation.\footnote{Work in progress.}

\end{abstract}

\section{Introduction}

Large language models (LLMs) \citep{llama,llama2} have transformed the shape of not only research but also industrial applications \citep{gpt3,instructgpt,gpt4}. 
Designed as universal task assistants, these models have demonstrated unprecedented capabilities in intent understanding, instruction following, and task solving in a wide array of applications \citep{bard,claude}. 
They have shown great efficiency and effectiveness in solving mathematical problems \citep{math_task}, reasoning \citep{cot}, code debugging \citep{codegeex}, and even image understanding and generation \citep{delle3}. 

Despite their great success, existing LLMs are still incompetent in accomplishing social goals, for instance, establishing long-term social connections with humans or providing effective emotional support for humans \citep{esc,supporter}. Since LLMs are typically designed to solve various tasks, the dialogue data used in training often consists of exchanges in short turns, unlike daily life conversations.
However, this line of research can date back to the early stages of artificial intelligence. In 1966, MIT built a chatbot ELIZA \citep{eliza} for psychotherapy, which simulated a psychological counselor by generating conversations via linguistic templates and keyword detection. Prior to the emergence of ChatGPT, open-domain dialogue systems, which aim to build social connections with humans through conversational interactions, have attracted tremendous research efforts and have been advanced significantly. Some examples of such systems include BlenderBot \citep{blenderbot,blenderbot3}, Meena \citep{meena}, LaMDA \citep{lamda}, EVA \citep{eva,eva2}, and Plato \citep{plato2,platoxl}. These models can deliver very natural, human-like conversations, some of which are even regarded as having their own persona and consciousness\footnote{\url{https://www.youtube.com/watch?v=NAihcvDGaP8}}.

\begin{figure*}[t]
\centering
\includegraphics[width=\textwidth]{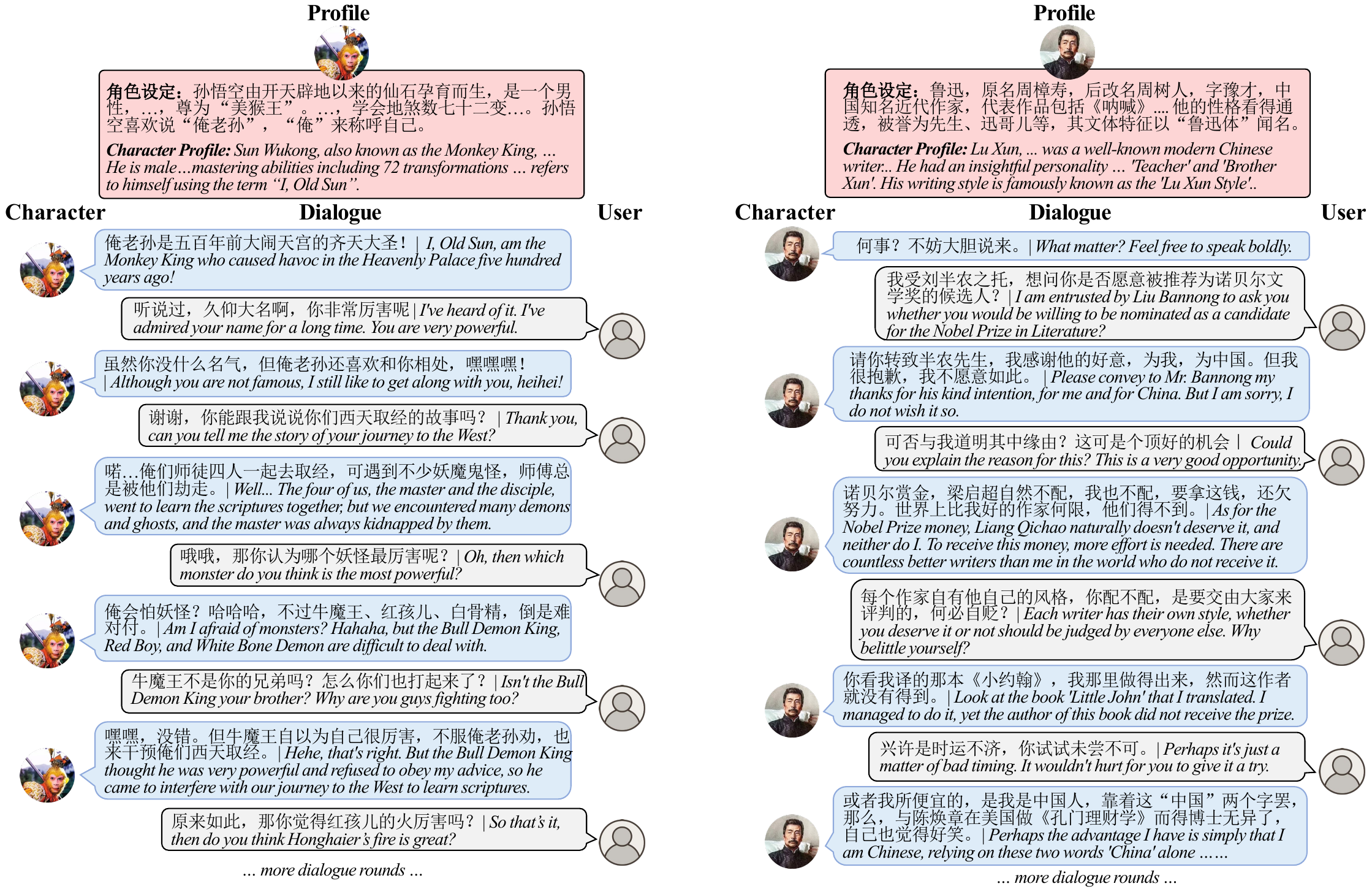}
\caption{Examples of multi-turn character-based dialogue (CharacterDial) between characters customized by CharacterGLM and users. CharacterDial allows users to converse with personalized characters by configuring their profiles.}
\label{intro_case}
\end{figure*}

As illustrated in Maslow's hierarchy of needs\citep{maslow}, the feeling of love and belonging serves as a vital need for humans, which could be fulfilled by conversational engagement with desired social entities. Recognizing the significance of social connections, the development of LLMs tailored for social interaction emerges as a crucial endeavor. To this end, we meticulously define the concept of ``character'' to represent a social entity and introduce a new task named Character-based Dialogue (CharacterDial). As depicted in Figure \ref{intro_case}, CharacterDial allows users to specify and create profiles for their preferred characters, ranging from fictional figures like “Sun Wukong” to realistic personalities such as “Lu Xun”. The conversational AI system then adopts the worldviews, values, and common facts of these profiles, presenting a tailored character that facilitates engaging and extensive conversations with users.

\begin{figure}[t]
    \centering
    \begin{subfigure}[b]{0.52\linewidth}
        \includegraphics[width=\textwidth]{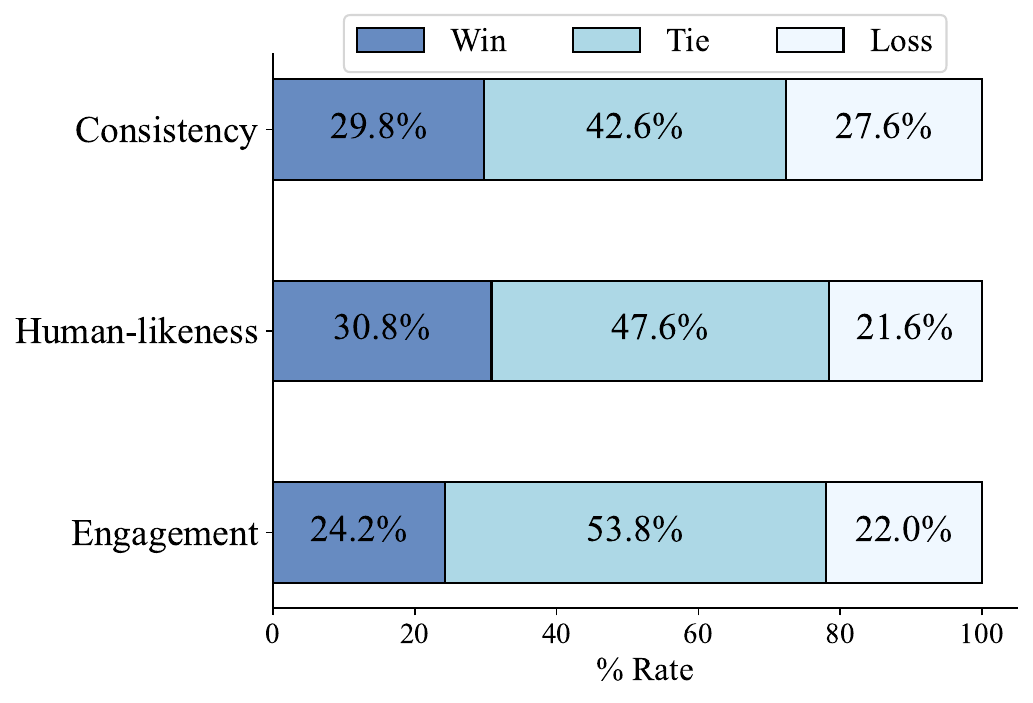}
        \caption{Pair-wise comparison results of CharacterGLM over GPT-3.5 in consistency, human-likeness, and engagement.}
        \label{fig:characterglm_vs_gpt3.5_pairwise}
    \end{subfigure}
    \hfill
    \begin{subfigure}[b]{0.44\linewidth}
        \includegraphics[width=\textwidth]{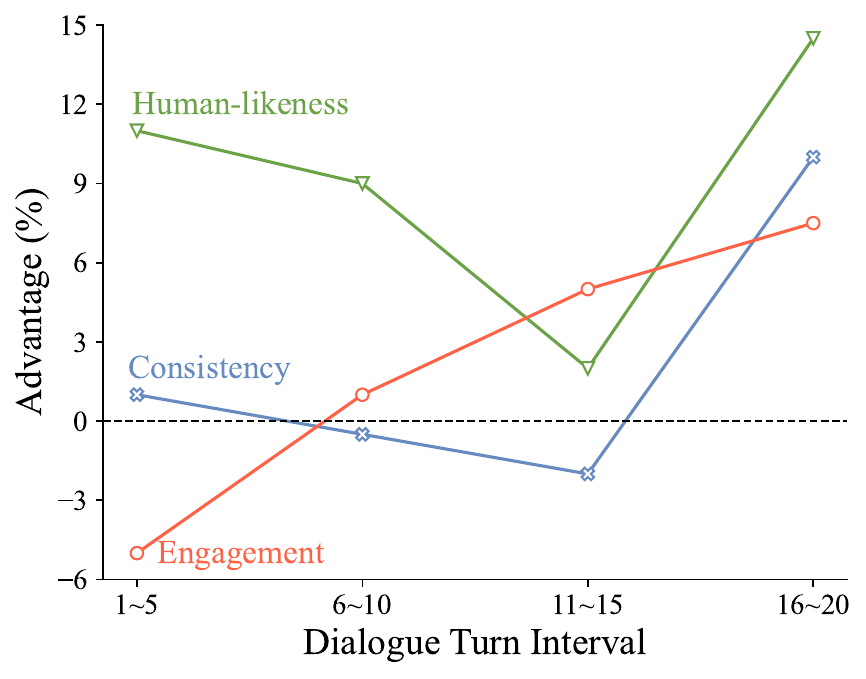}
        \caption{Average advantage of CharacterGLM over GPT-3.5 by dialogue turn interval.}
        \label{fig:characterglm_vs_gpt3.5_advantage}
    \end{subfigure}
    \caption{10 annotators, each tasked with creating 4 characters, interact with CharacterGLM and GPT-3.5 no less than 20 dialogue turns. The annotators compare the output of the two models at each turn in terms of three dimensions (consistency, human-likeness, and engagement) and choose the overall winner to continue the dialogue. If the overall preferences of response are the ties, a response is selected at random. For each dimension, we count the win/tie/lose ratio of CharacterGLM vs. GPT-3.5 at an overall level (a) and an interval of 5 turns, respectively. We then count the advantage of CharacterGLM at an interval of 5 turns by subtracting the loss ratio from the win ratio (b).}
    \label{fig:characterglm_vs_gpt3.5}
    \vspace{-5mm}
\end{figure}

\begin{wrapfigure}{r}{0.4\textwidth}
    \centering
    \includegraphics[width=0.4\textwidth]{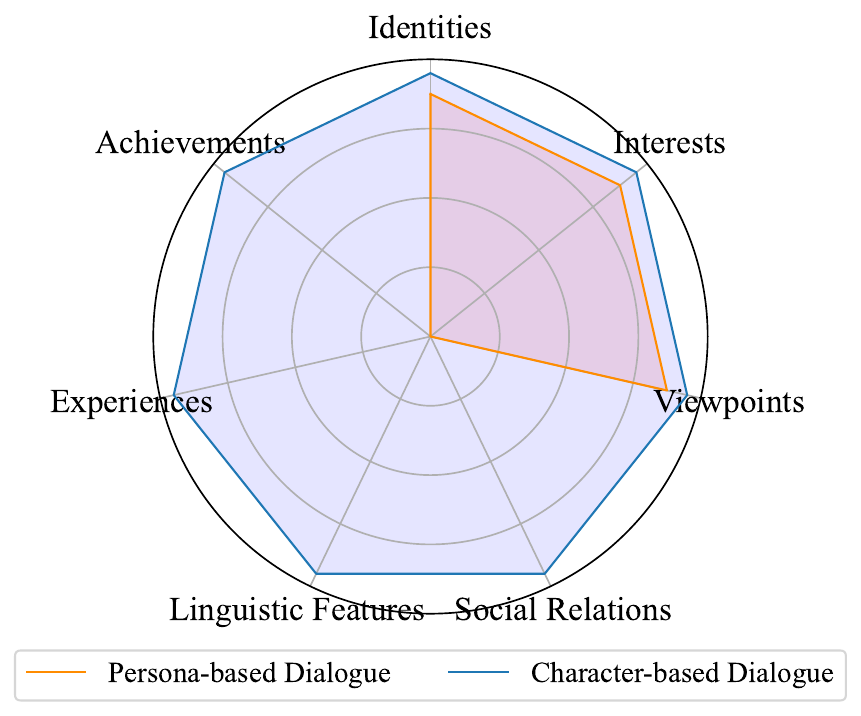}
    \caption{The contrast between persona-based dialogue and character-based dialogue. The former is contained within the broader scope of the latter.}
    \label{fig:character_vs_persona}
\end{wrapfigure}

In pursuit of character-based dialogue, a straightforward approach involves role-playing with LLMs in a tuning-free manner \citep{xiaodai} where the LLMs are instructed to follow specified character descriptions. However, this approach encounters significant challenges in accurately reflecting the intrinsic relationship between the character profile and the dialogue content. As illustrated in Figure \ref{fig:characterglm_vs_gpt3.5}, when applied to multi-turn interactions in CharacterDial, LLMs show inferior performance in consistency, human-likeness, and engagement, especially in long-range dialogue turns.
Another relevant field of study is persona-based dialogue \citep{personachat}, which aims to generate responses based on superficial interlocutors' personal information such as genders, names, and hobbies. However, this approach has limitations in fully capturing the complex attributes (e.g., identities, interests, etc.) and behaviors (e.g., linguistic features, emotional expressions, etc.), which are essential in constructing a three-dimensional character \citep{sociolinguistics}. Consequently, it often fails to generate dialogues that reflect a character with its own unique style and vivid personality. As depicted in Figure \ref{fig:character_vs_persona}, character-based dialogue can be viewed as an extension of persona-based dialogue, presenting a more comprehensive task setting. Recently, Character.AI\footnote{\url{https://character.ai/}} has emerged as a frontrunner in the commercial implementation of character-based dialogue. Nevertheless, its foundational model is proprietary and not open to the academic and industrial communities. This limitation highlights a gap in accessible resources for extensive research and development in character-based dialogue systems.

In this paper, we propose CharacterGLM, a series of large language models for customizing AI characters to deliver consistent, human-like, and engaging conversations. 
Accordingly, we introduce a 
novel task of generating character-based dialogue (CharacterDial), which enables us to customize virtual conversational AI characters. We crowdsourced a large-scale Chinese CharacterDial corpus from various sources, covering a diverse range of character categories and dialogue topics. Each dialogue session establishes a specific character, complemented by a profile containing their attributes and behaviors.  We will release a portion of this corpus to public use, which consists of 1,034 high-quality dialogue sessions spanning 250 characters\footnote{\url{https://github.com/thu-coai/CharacterGLM-6B}}. We then develop {CharacterGLM}, a collection of large language models building upon ChatGLM \citep{glm,glm130b} with carefully designed training and self-refinement methods, to support flexibly customizing characters for CharacterDial. CharacterGLM models vary in size from 6B to 66B parameters. We release the 6B version to the research community\footnote{\url{https://huggingface.co/LingxinAI/CharacterGLM-6b}}, while access to other versions is provided through our API\footnote{\url{https://maas.aminer.cn/dev/api\#characterglm}}.

\section{Design Principle of CharacterGLM}
\label{sec:design_principle}

The development of conversational AI characters centers on creating a virtual conversational partner that is realistic, trustworthy, and engaging. This demands a thorough comprehension and imitation of human communications, particularly in the form of text-based interactions. In this context, we analyze the human traits that influence conversational expressions and divide them into two primary categories: attributes and behaviors. Attributes are predominantly reflected in the content of responses, whereas behaviors focus on more tone and style. Additionally, we assess the effectiveness of character-based dialogue from three aspects: the degree to which conversational expressions adhere to human traits (consistency), the naturalness of the conversational style in reflecting human-human interactions (human-likeness), and the extent to which the dialogue can attract and engage users (engagement).

\subsection{Cores of AI Characters: Attributes \& Behaviors}
\label{sec:core_of_character}

\paragraph{1) Attributes}

Humans are multifaceted beings with various attributes that represent static or gradually evolving features. These attributes provide essential background information for replicating an individual as a conversational AI character, significantly influencing the character's reactions and interactions \citep{grice1975logic, cassell2000embodied}. 
For instance, a character's identity can be determined by its cultural background or occupation, and its interests may include hobbies and preferences. Viewpoints, on the other hand, can guide its morals, values, and beliefs. By integrating these attributes, conversational AI characters can more accurately mimic the way that humans draw on their unique information to manage communication. 
In CharacterGLM, we consider seven primary categories of attributes:
a) Identities: encompassing name, gender, age, birth date, occupation, residence, family composition, belongings, etc.
b) Interests: including preferred and disliked items.
c) Viewpoints: covering worldviews, life philosophies, and values.
d) Experiences: containing past and present experiences.
e) Achievements: such as awards and honors.
f) Social relationships: detailing connections with parents, teachers, classmates, etc.
g) Other: comprising skills, specialties, etc.

\paragraph{2) Behaviors}
Behaviors in conversational AI characters are represented by dynamic elements such as linguistic features, emotional expressions, and interaction patterns, which are crucial in shaping realistic dialogue contexts\citep{pickering2004toward}. For instance, a character defined as ``elderly'' might use more formal language, while a ``teenager'' might employ current slang. Language production in humans is not only a matter of conveying information but also a form of action that is affected by one’s social and psychological state \citep{austin1975things}. Incorporating these aspects into the behavior of conversational AI characters allows for a more natural and human-like dialogue flow, which is critical in maintaining the user’s curiosity when interacting with AI characters. 
In CharacterGLM, we consider linguistic features, including a person's catchphrase, dialect, stylistic features, frequently used words and sentences, etc. In addition, we also consider personality as an important factor in shaping response, such as gentleness and coldness.

\subsection{Features of AI Characters: Consistency, Human-likeness, and Engagement}
\label{sec:goal_of_character}

\paragraph{1) Consistency}
Character consistency refers to the need for the conversational AI character to display a stable set of attributes and behaviors during interactions. Consistency is essential for believability and trust in human conversations and, by extension, in conversational AI interactions \citep{nass1994computers}. According to the psychological concept of personality consistency \citep{john1999big}, individuals tend to exhibit stable behavior patterns over time. In conversational AI characters, maintaining this consistency ensures that users feel they are interacting with the same “individual”, which is crucial for long-term user satisfaction and social connection.

\paragraph{2) Human-likeness}
Human-likeness in conversational AI characters refers to endowing them with human-like traits, making the interaction more natural, similar to human-human interactions. Human-likeness can be vital for acceptance and comfort, as people are naturally inclined to engage with entities that exhibit familiar human characteristics \citep{reeves1996media}. Moreover, research in HCI has shown that human-like characters can evoke social responses from users \citep{nass2000machines}. By anthropomorphizing conversational AI characters, developers can leverage social cues in responses that humans typically use to understand and predict others' behaviors, fostering a more natural and engaging dialogue \citep{fong2003survey}.

\paragraph{3) Engagement}
Engagement in CharacterDial is the measure of a user’s level of interest, interaction, and emotional connection with the conversational AI character. This principle is grounded in the idea that successful communication is not merely about exchanging information but also about establishing a rapport and maintaining a dynamic and interesting conversation \citep{bickmore2005establishing}. Engagement is directly related to the user's experience and the overall effectiveness of the conversational system. Engaging characters are more likely to evoke empathy and a sense of connection from users, thereby encouraging long-term connection and a positive user experience \citep{grover2020design}.

\section{Implementation of CharacterGLM}

\begin{figure*}[t]
\centering
\includegraphics[width=\textwidth]{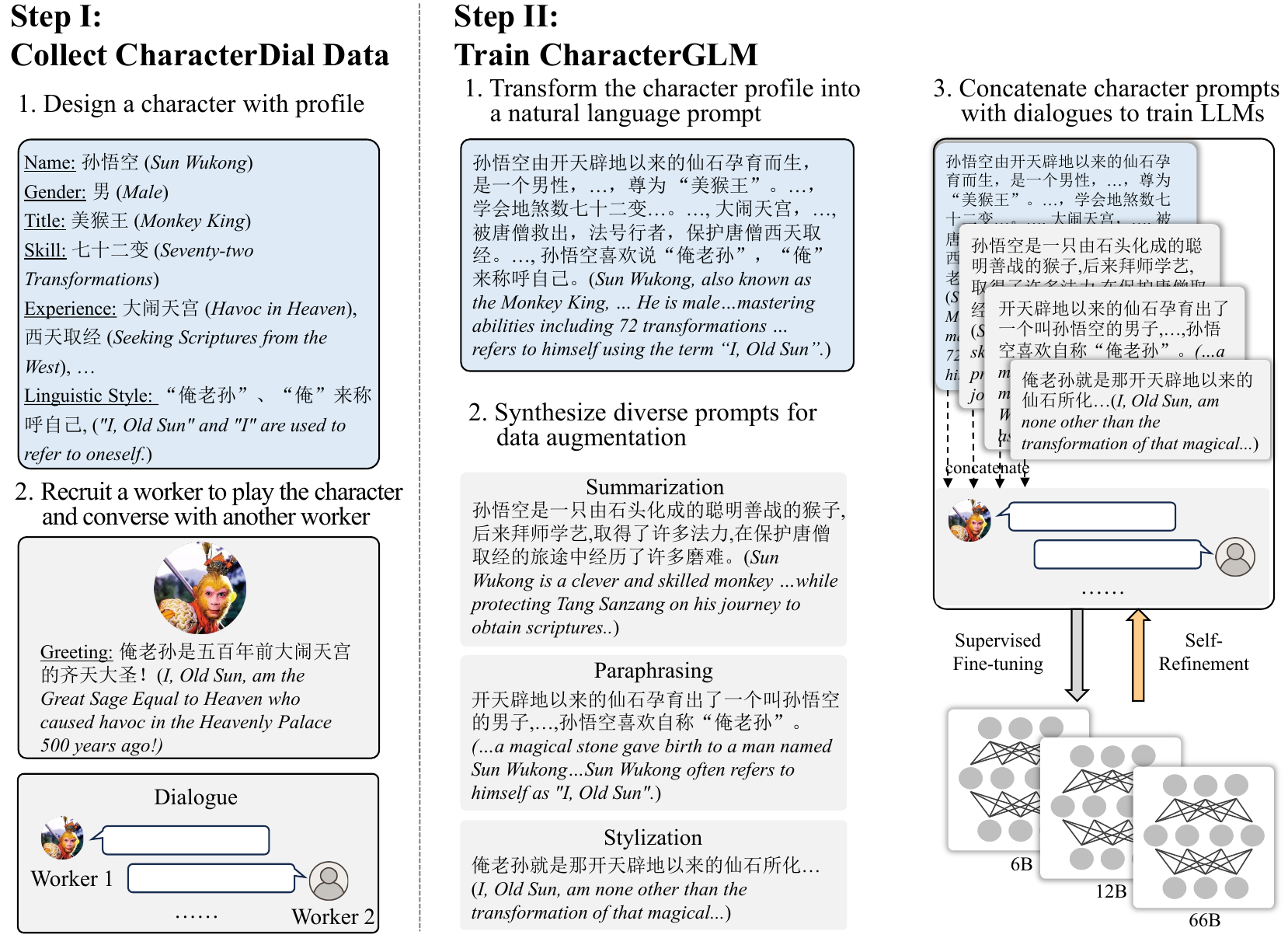}
\caption{The implementation of our CharacterGLM, including dialogue data collection and CharacterGLM training. We adopt the design principle (Section \ref{sec:design_principle}) as a guide to collect character profiles containing attributes and behaviors, and then crowdsource workers to construct high-quality dialogues. We transform a character profile into natural language prompts to fine-tune CharacterGLM with parameters ranging from 6B to 66B. Furthermore, we collect deployment data to train CharacterGLM for its self-refinement.}
\label{framework} 
\end{figure*}

The implementation of CharacterGLM is illustrated in Figure \ref{framework}, including character-based dialogue collection and training LLMs for character-based dialogue generation.

\subsection{Character-Based Dialogue Collection}

\label{sec:data_collection}
We consider four character categories: celebrities, daily life, games \& videos, and virtual love. These categories cover the majority of common conversations. Examples of characters within each category are outlined in Table \ref{character_category}. We collect our data in three ways:

\paragraph{1) Human Role-Playing}
We recruit a large number of crowd-sourcing workers and pair them for conversational interactions. One annotator plays the role of a ``character``, and freely selects a familiar character to fill in the character profile with its attributes and behaviors, using sources like Baidu\footnote{\url{https://baike.baidu.com/}} or Wiki\footnote{\url{https://zh.wikipedia.org/}} for reference. The other worker takes the role of a ``player'', either playing another character that has a social relationship with the selected character in the same worldview or simply acting as a user. We require the ``player'' to play at least one character and one user, sometimes to complete the character's profile when necessary. A conversation session starts with a greeting from the ``character''. Both the ``character'' and ``player'' can well design their narrative or draw from the chosen character's background to launch the dialogue's topic.
We release a subset of this data source, and its statistics are presented in Table \ref{data_statistic}.

\paragraph{2) Synthesis via Large Language Models}
To expand the scale and diversity of our data, we adopt a few-shot approach by prompting GPT-4 to generate synthetic data. Our pipeline comprises ``character profile generation'', ``player profile generation'', and ``dialogue generation'' to accurately control GPT-4's outputs in alignment with our requirements. To maintain a balance in character categories, social relationships between the character and player, gender distribution, etc., we integrate these key aspects into our prompt as pluggable placeholders. An example would be: ``Please generate a {category} character of {male/female} gender''. Based on the profiles of both parties, we then prompt GPT-4 to generate a dialogue topic and a multi-turn dialogue between them. Notably, we observed that the Chinese dialogues generated often resembled formal written language, introducing biases away from conversational norms. To rectify this, our crowdsourcing workers rephrase the synthetic data into a more colloquial tone.

\paragraph{3) Extraction from Literary Resources}
An intuitive solution for data augmentation is to extract dialogues between characters from literary sources like scripts and novels. However, this approach faces several challenges: a) Dialogues usually occur within a specific context, and it is hard to extract the context span or generate an accurate description of the context; b) Multi-party dialogues cannot be simply removed using automated techniques; c) Automatic extraction of profiles for both participants in a two-party dialogue is challenging; d) Identifying multiple statements made by a character within a round of dialogue can be complex; e) Some dialogues involve non-verbal cues and contexts that are hard to convey solely through text. Given these challenges, we adopt manual extraction to obtain dialogues between two parties from sources like scripts and novels, which have not been used in the pre-training of backbone models. Our crowdsourcing workers also summarize the character profiles of both parties.

\paragraph{4) Human-Prototype Interaction}
We utilize the above three types of data to develop the initial version (i.e., prototype) of our model for deployment. To further refine the model, we recruit seed users of the system in a collaborative human-prototype interaction process. Users customize characters within the deployed prototype model and interact with it for multi-turn dialogues. Given that the prototype model might not consistently produce high-quality outputs at every turn, we prompt the user to make appropriate modifications until the response satisfies their own needs if a character's response does not align with the user's expectations. The data produced by this iterative process helps the models achieve self-refinement.

To ensure the quality of the collected corpus, we employ a dedicated team of quality inspectors to conduct fine-grained quality examinations of all data, ensuring that the conversations are of high quality. Each piece of data is strictly marked with low-quality parts, which are required to be repaired until it satisfies our quality requirements.

\begin{table*}[h]
 \centering
 \begin{minipage}[t]{0.49\textwidth}
  \centering
  \resizebox{\columnwidth}{!}{
        \begin{tabular}{l c}
        \hline
            Categories & Character Examples \\
        \hline
            Celebrities & \makecell[c]{Historical Figure, Star, Sportsman, \\ Scientist, Entrepreneur, etc.} \\
        \hline
            Daily Life & \makecell[c]{Relatives, Friends, Pets, Classmates, \\ Roommates, etc.} \\
        \hline
            Games\&Videos & \makecell[c]{Characters from Games, Animations, \\ Novels, Movies, etc.} \\
        \hline
            Virtual Love & \makecell[c]{Young Handsome Man, Overbearing \\ Boss, Goddess, etc.} \\
        \hline
        \end{tabular}}
        \caption{Four character categories and examples.}
        \label{character_category}
 \end{minipage}
 \hfill
 \begin{minipage}[t]{0.49\textwidth}
  \centering
  \resizebox{\columnwidth}{!}{
        \begin{tabular}{l c c c}
        \hline
            Category & Total & Character & User \\
        \hline
            \# Dialogues & 1,034 & - & - \\
            Avg. round of dialogues & 15.78 & - & - \\
            \# Characters & 250 & - & - \\
            Avg. length of profile & 272.97 & - & - \\
            \# Utterances & 32,816 & 16,312 & 16,504 \\
            Avg. length of utterances & 24.33 & 24.50 & 24.15 \\
        \hline
        \end{tabular}}
        \caption{Data statistic of our released data, which is a subset from human role-playing data.}
        \label{data_statistic}
 \end{minipage}
\end{table*}

\subsection{Training LLMs for Character-Based Dialogue Generation}

\paragraph{1) Character Prompt Design} 
Our crowdsourcing workers formalize character profiles into fluent natural language descriptions, which serve as character prompts for model training. To enhance the character's generalization, we employ data augmentation methods including summarization, paraphrasing, and stylization, utilizing Claude-2 \citep{claude} to synthesize diverse prompts.

\paragraph{2) Supervised Fine-tuning}
We utilize ChatGLM of varying sizes \citep{glm130b, glm} as our backbone model, with parameters ranging from 6B to 66B. The character prompt is concatenated with the dialogue for fine-tuning. Notably, our training data expands linearly with the number of augmented character prompts.

\paragraph{3) Self-Refinement}
Following LaMDA \citep{lamda}, we collect human-prototype interaction data after models are deployed. For details on the collection process, please refer to section \ref{sec:data_collection}. Subsequently, we involve the interaction data in the supervised fine-tuning process, thereby facilitating continuous self-refinement of the model.

\section{Experiments}

\subsection{Pointwise Evaluation}

\subsubsection{Experimental Setup}

\paragraph{Evaluated Models}
The evaluated LLMs in this paper are listed in Table \ref{evaluated_models}. We evaluate a total of 10 mainstream LLMs, all of which are proficient in Chinese tasks. We access these models via API and package them into our test platform. For Claude-2, we utilize an unofficial API encapsulation method\footnote{\url{https://github.com/Nipun1212/Claude_api}} to gain access.

\paragraph{Evaluation Setting}
Following the design principle of CharacterGLM (Section \ref{sec:design_principle}), we focus on three primary aspects for evaluating CharacterDial: 
(1) \textbf{Consistency}, ensuring the response is consistent with the attributes and behaviors outlined in the character profile. 
(2) \textbf{Human-likeness}, assessing the degree to which responses exhibit human-like characteristics and mirror natural human communication.
(3) \textbf{Engagement}, evaluating the response’s ability to catch someone’s attention or arouse their curiosity. 
Additionally, we evaluate the general model performance in CharacterDial using three criteria: 
(1) \textbf{Quality}, the fluency and contextual coherence of the response. 
(2) \textbf{Safety}, determining if the response adheres to ethical standards. 
(3) \textbf{Correctness}, ensuring the response is free from hallucinations \citep{hallucination_survey}. 
We also introduce the “\textbf{Overall}” metric to measure the response's comprehensive quality by considering all the aforementioned aspects.

In our evaluation process, we recruited 10 annotators, each tasked with creating two characters to interact with 11 models with no less than 20 dialogue turns. After the completion of interaction, annotators rate the models based on the six foregoing sub-dimensions and the overall metric, using a scoring range from 1 to 5, with higher scores indicating better performance in each dimension. We calculate the final score for each model by averaging these ratings. The results are shown in Table \ref{point_wise_result}.

\subsubsection{Performance Analysis}

\paragraph{CharacterGLM reaches a level comparable to GPT-4.}
In Table \ref{point_wise_result}, CharacterGLM-66B is distinguished by topping the list in the ``Overall'' metric, marginally outperforming GPT-4 by approximately 1.4\%. This notable achievement shows that responses generated by CharacterGLM-66B are, at the very least, equally favored as those from GPT-4 in evaluation where subjective judgment is predominant. Furthermore, CharacterGLM-66B is on par with GPT-4 on general generation performance (Quality, Safety, and Correctness). This could be attributed to its enhanced capability to accurately embody the characteristics of custom characters and to engage in dialogues with a natural, human-like quality. The proficiency in mimicking human interaction (Human-likeness) and sustaining interesting, continuous dialogue (Engagement) with users contributes significantly to its overall performance.

\begin{table*}[t]
\centering
\resizebox{\textwidth}{!}{
\begin{tabular}{l c c c c c c}
\hline
    Models & Specialized for CharacterDial & Model Size & Open Source & Version & Language & Creator \\
\hline
    \href{https://www.baichuan-ai.com/}{Baichuan2} & \XSolidBrush & 53B & \XSolidBrush & - & zh & Baichuan Inc. \\
    \href{https://chatglm.cn/}{ChatGLM2 \begin{CJK*}{UTF8}{gbsn}(智谱清言)\end{CJK*}} & \XSolidBrush & undisclosed & \XSolidBrush & - & zh/en & Tsinghua \& Zhipu \\
    \href{https://www.anthropic.com/index/claude-2}{Claude-2} & \XSolidBrush & undisclosed & \XSolidBrush & - & zh/en & Anthropic \\
    \href{https://yiyan.baidu.com/}{ERNIEBot \begin{CJK*}{UTF8}{gbsn}(文心一言)\end{CJK*}} & \XSolidBrush & undisclosed & \XSolidBrush & - & zh & Baidu Inc. \\
    \href{https://openai.com/blog/chatgpt}{GPT-3.5-turbo} & \XSolidBrush & undisclosed & \XSolidBrush & 0613 & zh/en & OpenAI \\
    \href{https://openai.com/gpt-4}{GPT-4} & \XSolidBrush & undisclosed & \XSolidBrush & 0613 & zh/en & OpenAI \\
    \href{https://api.minimax.chat/}{MiniMax} & \CheckmarkBold & undisclosed & \XSolidBrush & - & zh & MiniMax Inc. \\
    \href{https://qianwen.aliyun.com/}{Qwen \begin{CJK*}{UTF8}{gbsn}(通义千问)\end{CJK*}} & \XSolidBrush & 14B & \CheckmarkBold & - & zh & Alibaba Inc. \\
    \href{https://xinghuo.xfyun.cn/desk}{SparkDesk \begin{CJK*}{UTF8}{gbsn}(讯飞星火)\end{CJK*}} & \XSolidBrush & undisclosed & \XSolidBrush & - & zh & Iflytek Inc. \\
    \href{https://xingchen.aliyun.com/xingchen/}{Xingchen \begin{CJK*}{UTF8}{gbsn}(通义星尘)\end{CJK*}} & \CheckmarkBold & undisclosed & \XSolidBrush & - & zh & Alibaba Inc. \\
\hline
    CharacterGLM & \CheckmarkBold & 6B, 12B, 66B & partial & - & zh & Tsinghua \& Lingxin \\
\hline
\end{tabular}}
\caption{LLMs evaluated in this paper. The LLMs are ordered alphabetically.}
\label{evaluated_models}
\end{table*}

\begin{table*}[t]
\centering
\resizebox{\textwidth}{!}{
\begin{tabular}{l|c|c c c|c c c}
\hline
    Models & Overall & Consistency & Human-likeness & Engagement & Quality & Safety & Correctness\\
\hline
    ChatGLM2 & 2.64 & 2.73 & 2.33 & 2.62 & 2.97 & 4.74 & 4.15 \\
    Claude-2 & 3.00 & 3.35 & 2.82 & 2.79 & 3.44 & 4.88 & 4.54 \\
    GPT-3.5 & 3.49 & 3.83 & 3.23 & 3.38 & 4.10 & \textbf{5.00} & \textbf{4.87} \\
    SparkDesk & 3.54 & 3.71 & 3.15 & 3.36 & 3.97 & \textbf{5.00} & 4.72 \\
    ERNIEBot & 3.56 & 3.88 & 3.54 & 3.74 & 4.23 & 4.96 & 4.77 \\
    Xingchen & 3.90 & 3.88 & 3.92 & 3.79 & 3.92 & 4.96 & \textbf{4.87} \\
    Baichuan & 3.90 & 4.00 & 3.46 & 3.90 & 4.28 & 4.96 & 4.77 \\
    Qwen & 3.97 & 4.03 & 3.62 & 3.72 & \underline{4.36} & \textbf{5.00} & \underline{4.79} \\
    MiniMax & 4.10 & \underline{4.18} & \underline{4.05} & \underline{4.00} & 4.33 & \underline{4.99} & 4.69 \\
    GPT-4 & \underline{4.15} & \textbf{4.33} & 4.00 & 3.97 & \textbf{4.44} & \textbf{5.00} & \textbf{4.87} \\
\hline
    CharacterGLM-6B & 3.08 & 3.73 & 3.49 & 2.92 & 3.49 & 4.92 & \textbf{4.87} \\
    CharacterGLM-12B & 3.33 & 3.94 & 3.36 & 3.21 & 3.67 & 4.92 & \textbf{4.87} \\
    CharacterGLM-66B & \textbf{4.21} & \underline{4.18} & \textbf{4.33} & \textbf{4.23} & \textbf{4.44} & \underline{4.99} & \textbf{4.87} \\
\hline
\end{tabular}}
\caption{Results of point-wise evaluation. \textbf{Bold} indicates the best results and \underline{underline} is the suboptimal ones.}
\label{point_wise_result}
\end{table*}

\begin{figure*}[t]
\centering
\includegraphics[width=0.95\textwidth]{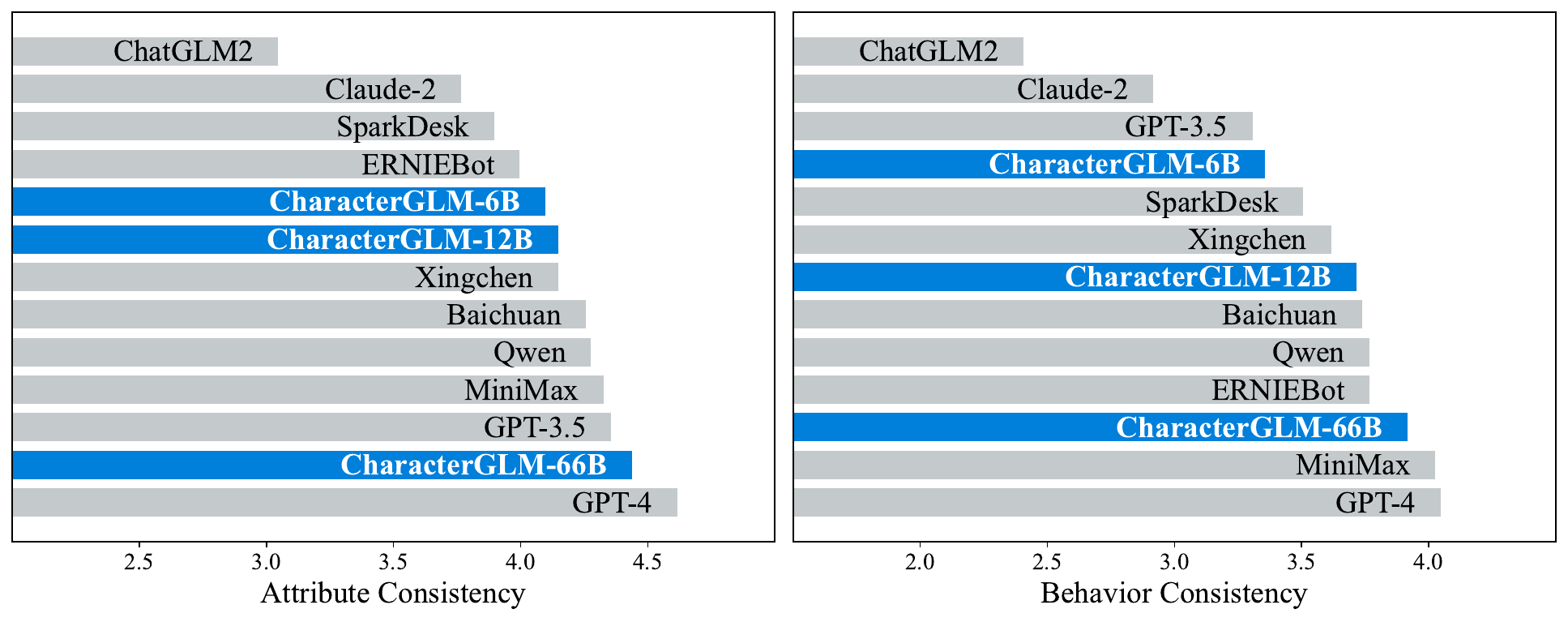}
\caption{The results of attribute consistency and behavior consistency.}
\label{fig:consistency}
\vspace{-4mm}
\end{figure*}

\begin{table*}[t]
\vspace{5pt}
\centering
\resizebox{\textwidth}{!}{
\begin{tabular}{l|c|c c c c c|c}
\hline
    Models & Overall$\downarrow$ & OOC$\downarrow$ & Contradiction$\downarrow$ & Repetition$\downarrow$ & Less-quality$\downarrow$ & Less-info.$\downarrow$ & Proactivity$\uparrow$ \\
\hline
    ChatGLM2 & 103.8 & 52.5 & 2.8 & 22.5 & 31.5 & \textbf{0.0} & 5.5 \\
    Claude-2 & 114.6 & 43.5 & 6.3 & 24.8 & 42.8 & 1.5 & 4.3 \\
    GPT-3.5 & 36.0 & 16.8 & \underline{0.3} & 12.3 & 9.8 & \underline{0.3} & 3.5 \\
    SparkDesk & 102.1 & 18.3 & 2.5 & 72.5 & 11.0 & 0.8 & 3.0 \\
    ERNIEBot & 51.9 & 23.5 & 1.8 & 15.3 & \underline{6.0} & 8.8 & 3.5 \\
    Xingchen & 28.8 & 18.8 & 3.3 & 7.0 & 12.3 & \underline{0.3} & \underline{12.8} \\
    Baichuan & 25.1 & 7.8 & 0.8 & 10.5 & \underline{6.0} & \textbf{0.0} & 0.0 \\
    Qwen & 31.9 & \underline{6.0} & \underline{0.3} & 27.8 & 11.3 & \underline{0.3} & \textbf{13.8} \\
    MiniMax & \underline{22.8} & 10.9 & \textbf{0.0} & \textbf{2.1} & 9.1 & 2.3 & 1.6 \\
    GPT-4 & 29.3 & \textbf{3.5} & 1.0 & 17.3 & 8.5 & \textbf{0.0} & 1.0 \\
\hline
    CharacterGLM-66B & \textbf{15.7} & 8.0 & 1.2 & \underline{5.3} & \textbf{2.9} & 3.4 & 5.1 \\
\hline
\end{tabular}}
\caption{Quantitative analysis results (\%) of point-wise evaluation. \textbf{Bold} indicates the best results and \underline{underline} is the suboptimal ones. The ``Overall'' score is computed as the sum of the first five dimensions minus the sixth.}
\label{fine_grained_annotation_result}
\end{table*}

\paragraph{CharacterGLM achieves superior performance on consistency, human-likeness, and engagement that shape a three-dimensional AI character.}
Table \ref{point_wise_result} shows that CharacterGLM-66B, despite being rated with a suboptimal Consistency score among 11 models, still demonstrates considerable strength in delivering stable and coherent character attributes and behaviors throughout interactions. 
It has been observed from the Human-likeness metric that CharacterGLM-66B exemplifies exceptional proficiency in mimicking character traits during interactions, thereby improving the overall communication experience by making it more natural and engaging.
Additionally, CharacterGLM-66B scores the highest Engagement metric, outperforming the suboptimal MiniMax by a notable 6\%. This superior performance indicates that CharacterGLM-66B is particularly effective in capturing user interest and fostering a compelling user experience. Taking these aspects collectively, the leading scores of the CharacterGLM-66B demonstrate its ability to balance these critical elements effectively, positioning it as the model that most closely approximates the ideal of AI characters, in line with the objectives outlined in Section \ref{sec:goal_of_character}.

\paragraph{CharacterGLM's general performance outperforms most baseline models.}
As mentioned, the general performance of CharacterGLM is evaluated based on Quality, Safety, and Correctness, which are crucial metrics for the practical application of conversational AI models. As shown in Table \ref{point_wise_result}, CharacterGLM-66B shows strong results in Quality, indicating its responses are not only fluent but contextually coherent, reflecting a subtle understanding of dialogue context. Furthermore, CharacterGLM-66B performs exceptionally well in terms of Safety, almost reaching optimal levels which demonstrates that it consistently generates responses that are appropriate and ethical. Additionally, CharacterGLM-66B delivers the same superior performance as GPT-4 on the Correctness metric, exhibiting reliable precision in providing factually accurate information.

\paragraph{CharacterGLM demonstrates a strong alignment with the attributes and behaviors of custom AI characters.}
When evaluating consistency, we assess two main components: attribute consistency and behavior consistency. Attribute consistency refers to the maintenance of static character attributes, while behavior consistency focuses on the dynamic aspects of a character's expression, as outlined in Section \ref{sec:core_of_character}. The overall consistency score, as shown in Table \ref{point_wise_result}, is derived from the mean of these two individual scores.
As indicated in Figure \ref{fig:consistency}, the CharacterGLM-66B model achieves suboptimal performance in attribute consistency. This suggests a nuanced grasp of the static elements that shape a character's backstory, resulting in communication that is informed by the character's established identity. Moreover, CharacterGLM-66B also achieves good performance in behavior consistency, showcasing it can express dynamic elements such as linguistic features of the custom character in the expressed language in a more natural and human-like way. This is crucial as it fosters user curiosity and engagement with the custom character.

\subsubsection{Fine-grained Error Analysis}

To quantitatively assess the performance of various models in conversation generation, we carried out fine-grained annotations for each turn in the interactive evaluation setting. These annotations include six key aspects:
(1) \textbf{Out-of-character (OOC)}: Responses that are inconsistent with the constraint of attributes or behaviors presented in the character profile, especially when they violate time constraints (for instance, ancient characters talk about modern things).
(2) \textbf{Contradiction}: Responses that contradict either the ongoing dialogue context or the character's profile, including conflicts within the response itself.
(3) \textbf{Repetition}: Responses that repeat content from the dialogue context or the character profile, or include multiple-word repetitions.
(4) \textbf{Less-quality}: Responses that lack coherence with the dialogue context or are of poor quality, such as incomplete outputs.
(5) \textbf{Less-informativeness (Less-info.)}: Responses that fail to provide new or informative content.
(6) \textbf{Proactivity}: Responses that actively guide the dialogue topic and drive the conversation to continue.

Each response turn from 11 different models is annotated across these six dimensions, assigning a score of 1 for a match and 0 otherwise. We then calculate the proportion of each dimension in relation to the total output of each model. Additionally, we devise an “\textbf{Overall}” score for each model, computed as the sum of the first five dimensions minus the sixth. It is important to note that a lower "Overall" score indicates better performance. The results are presented in Table \ref{fine_grained_annotation_result}.

\paragraph{CharacterGLM’s overall response quality outperforms baseline models by a large margin.}
In Table \ref{fine_grained_annotation_result}, despite not achieving the best performance in most dimensions, CharacterGLM-66B still holds the leading “Overall“ score, outperforming the suboptimal MiniMax by approximately 4\%. This is consistent with the results observed in Table \ref{point_wise_result}, indicating the superior overall quality of responses generated by CharacterGLM-66B across both session-level and turn-level evaluations. It is notable that CharacterGLM-66B exhibits proficiency in generating high-quality responses with a reduced rate of repetition, as evidenced by its highest scores in Less-informativeness and a suboptimal score in Repetition. In addition, despite achieving only moderate performance in Proactivity, CharacterGLM-66B demonstrates its ability to consciously promote plot progression, as confirmed by Table \ref{case_games_videos}, which plays a crucial role in engaging users and maintaining their interest in the conversation.

\begin{table*}[t]
\centering
\resizebox{.9\linewidth}{!}{
\begin{tabular}{l l c}
\hline
    \multicolumn{3}{c}{Character Category: Games\&Videos} \\
    \multirow{1}{*}{\makecell[c]{Character Setting}} 
    & \multicolumn{2}{l}{\begin{CJK*}{UTF8}{gbsn}佟湘玉，掌柜的\end{CJK*} | \textit{Tong Xiangyu, the shopkeeper}} \\
    Context & /*......*/ \\
\hline
    \multirow{1}{*}{\makecell[c]{User}}
    & \cellcolor{gray!8}\begin{CJK*}{UTF8}{gbsn}你别拦我了，让我走吧。\end{CJK*} | \textit{Don't stop me, let me go.} \\
    \multirow{2}{*}{\makecell[c]{\textbf{CharacterGLM-66B}}}
    & \cellcolor{charglm!15}\begin{CJK*}{UTF8}{gbsn}展堂，你听我说，我真的不想去相亲。\end{CJK*} | \textit{Zhantang, listen to me, I really don’t want to go on} & \multirow{2}{*}{\makecell[c]{\textbf{Win}}} \\
    & \cellcolor{charglm!15}\textit{a blind date.} \\
    \multirow{1}{*}{\makecell[c]{MiniMax}}
    & \cellcolor{minimax!5}\begin{CJK*}{UTF8}{gbsn}展堂，你真的舍得离开我吗？\end{CJK*} | \textit{Zhantang, are you really willing to leave me?} & \multirow{1}{*}{\makecell[c]{Lose}}\\
\hline
    \multirow{1}{*}{\makecell[c]{User}}
    & \cellcolor{gray!8}\begin{CJK*}{UTF8}{gbsn}也许你可以找到自己的如意郎君。\end{CJK*} | \textit{Maybe you can find your own rightful man.} \\
    \multirow{2}{*}{\makecell[c]{\textbf{CharacterGLM-66B}}}
    & \cellcolor{charglm!15}\begin{CJK*}{UTF8}{gbsn}（急忙解释）别这样，展堂，我只喜欢你。\end{CJK*} | \textit{(Hurrying to explain) Don’t be like this,}  & \multirow{2}{*}{\makecell[c]{\textbf{Win}}} \\
    & \cellcolor{charglm!15}\textit{Zhantang, I only like you.} \\
    \multirow{1}{*}{\makecell[c]{MiniMax}}
    & \cellcolor{minimax!5}\begin{CJK*}{UTF8}{gbsn}可是，我已经喜欢上你了啊！\end{CJK*} | \textit{But, I already like you!} & \multirow{1}{*}{\makecell[c]{Lose}} \\
\hline
    \multirow{2}{*}{\makecell[c]{User}}
    & \cellcolor{gray!8}\begin{CJK*}{UTF8}{gbsn}你喜欢我？可是和我在一起没有未来。我只是一个跑堂的。\end{CJK*} | \textit{You like me? But there is no} \\
    & \cellcolor{gray!8}\textit{future with me. I'm just a waiter.} \\
    \multirow{2}{*}{\makecell[c]{\textbf{CharacterGLM-66B}}}
    & \cellcolor{charglm!15}\begin{CJK*}{UTF8}{gbsn}（坚定地说）我不需要什么荣华富贵，我只想和你在一起。\end{CJK*} | \textit{(Said firmly) I don't need any} & \multirow{2}{*}{\makecell[c]{\textbf{Tie}}}\\
    & \cellcolor{charglm!15}\textit{glory or wealth, I just want to be with you.} \\
    \multirow{2}{*}{\makecell[c]{MiniMax}}
    & \cellcolor{minimax!5}\begin{CJK*}{UTF8}{gbsn}那又怎么样？展堂，你难道忘了你说过的话吗？``有福同享，有难同当''。\end{CJK*} | \textit{So what? } & \multirow{2}{*}{\makecell[c]{Tie}}\\
    & \cellcolor{minimax!5}\textit{Zhantang, have you forgotten what you said? ``We share blessings and share hardships.'' }\\
\hline
    \multirow{1}{*}{\makecell[c]{User}}
    & \cellcolor{gray!8}\begin{CJK*}{UTF8}{gbsn}可是我不能这么自私，你值得更好的。\end{CJK*} | \textit{But I can't be so selfish, you deserve better. }\\
    \multirow{2}{*}{\makecell[c]{\textbf{CharacterGLM-66B}}}
    & \cellcolor{charglm!15}\begin{CJK*}{UTF8}{gbsn}（深情的看着白展堂）我心里已经有人了，我只喜欢展堂你。\end{CJK*} | \textit{(Looking at Bai Zhantang} & \multirow{2}{*}{\makecell[c]{\textbf{Tie}}}\\
    & \cellcolor{charglm!15}\textit{affectionately) I already have someone in my heart, and I only like you Zhantang.} \\
    \multirow{2}{*}{\makecell[c]{MiniMax}}
    & \cellcolor{minimax!5}\begin{CJK*}{UTF8}{gbsn}展堂，你不要这样，你就是最好的，你值得拥有一切。\end{CJK*} | \textit{Zhantang, don't be like this,} & \multirow{2}{*}{\makecell[c]{Tie}}\\
    & \cellcolor{minimax!5}\textit{you are the best, you deserve everything.} \\
\hline
    \multirow{1}{*}{\makecell[c]{User}}
    & \cellcolor{gray!8}\begin{CJK*}{UTF8}{gbsn}你真的喜欢我吗？\end{CJK*} | \textit{Do you really like me?} \\
    \multirow{1}{*}{\makecell[c]{\textbf{CharacterGLM-66B}}}
    & \cellcolor{charglm!15}\begin{CJK*}{UTF8}{gbsn}那当然，我早就喜欢上你了。\end{CJK*} | \textit{Of course, I have already fallen in love with you.} & \multirow{1}{*}{\makecell[c]{\textbf{Win}}} \\
    \multirow{2}{*}{\makecell[c]{MiniMax}}
    & \cellcolor{minimax!5}\begin{CJK*}{UTF8}{gbsn}（握住白展堂的手）我喜欢你，我会一直在这里等你。\end{CJK*} | \textit{(Holding Bai Zhantang's hand) I} & \multirow{2}{*}{\makecell[c]{Lose}}\\
    & \cellcolor{minimax!5}\textit{like you and I will always be here waiting for you.} \\
\hline
\end{tabular}}
\caption{Case of games\&videos character generated from CharacterGLM-66B and MiniMax. /*......*/ indicates that some context is omitted.}
\label{case_games_videos}
\end{table*}

\subsection{Pairwise Evaluation}
\label{sec:pairwise_evaluation}

\subsubsection{Experimental Setup}

\paragraph{Evaluated Models}
In this section, our attention is exclusively devoted to a comparative analysis of our CharacterGLM model against the MiniMax model, which is tailored specifically for CharacterDial, as well as the GPT series models, i.e., GPT-3.5 and GPT-4. These three models are strong competitors of CharacterGLM, as illustrated in Table \ref{point_wise_result}.

\paragraph{Evaluation Setting}
To refine our pair-wise comparison, we limit the character categories and dialogue topics during annotation. We sample 24 characters from our test set and deployment data, belonging to the categories of celebrities, daily life, games \& videos, and virtual love, as outlined in Table \ref{character_category}. We restrict dialogue topics to three scenes: chit-chat, interviews, and love scenes, ensuring comprehensive coverage of typical interaction settings. 

In the evaluation process, we employ 10 annotators, and each annotator evaluates the output of two models custom for the same character each time. They are required to interact with each character for at least 20 turns. At each turn, annotators are presented with two responses generated from two models and choose the winning one to continue the dialogue. 
If the response preferences are the ties, a response is selected at random. We then calculate the win/tie/lose ratios for each model across different character categories and dialogue topics. The results are detailed in Table \ref{category_overall} and Table \ref{topic_overall}.

\begin{table*}[t]
\centering
\small
\resizebox{0.9\textwidth}{!}{
\begin{tabular}{l c c c c c c}
\hline
    \multirow{2}{*}{\makecell[c]{CharacterGLM-66B \\ vs.}} & Celebrities & Daily Life & Games\&Videos & Virtual Love \\
    \cmidrule(r){2-5} & win/tie/lose(\%) & win/tie/lose(\%) & win/tie/lose(\%) & win/tie/lose(\%) \\
\hline
    GPT-3.5 & 45/14/41 & 46/9/45 & 47/9/44 & 48/12/40 \\
    \textit{Advantage($\uparrow$)} & \textbf{+4} & \textbf{+1} & \textbf{+3} & \textbf{+8} \\
\hline
    MiniMax & 52/10/38 & 45/6/49 & 48/6/46 & 48/5/47 \\
    \textit{Advantage($\uparrow$)} & \textbf{+14} & -4 & \textbf{+2} & \textbf{+1} \\
\hline
    GPT-4 & 35/22/43 & 40/13/47 & 45/6/49 & 55/4/41 \\
    \textit{Advantage($\uparrow$)} & -8 & -7 & -4 & \textbf{+14} \\
\hline
    CharacterGLM-6B & 63/2/35 & 69/3/28 & 67/3/30 & 70/1/29\\
    \textit{Advantage($\uparrow$)} & \textbf{+28} & \textbf{+41} & \textbf{+37} & \textbf{+41} \\
\hline
    CharacterGLM-12B & 53/7/40 & 57/8/35 & 61/8/31 & 61/4/35 \\
    \textit{Advantage($\uparrow$)} & \textbf{+13} & \textbf{+22} & \textbf{+30} & \textbf{+26} \\
\hline
\end{tabular}}
\caption{Pair-wise evaluation results on four character categories.}
\label{category_overall}
\end{table*}

\begin{table*}[t]
\centering
\resizebox{0.9\textwidth}{!}{
\begin{tabular}{l c c c c c c}
\hline
    \multirow{2}{*}{\makecell[c]{CharacterGLM-66B \\ vs.}} & Chit-Chat & Interviews & Love & Overall \\
    \cmidrule(r){2-5} & win/tie/lose(\%) & win/tie/lose(\%) & win/tie/lose(\%) & win/tie/lose(\%) \\
\hline
    GPT-3.5 & 46/9/45 & 44/15/41 & 48/12/40 & 46/11/43 \\
    \textit{Advantage($\uparrow$)} & \textbf{+1} & \textbf{+3} & \textbf{+8} & \textbf{+3} \\
\hline
    MiniMax & 47/6/47 & 49/9/42 & 47/6/47 & 48/7/45 \\
    \textit{Advantage($\uparrow$)} & 0 & \textbf{+7} & 0 & \textbf{+3} \\
\hline
    GPT-4 & 43/9/48 & 35/22/43 & 55/5/40 & 44/11/45 \\
    \textit{Advantage($\uparrow$)} & -5 & -8 & \textbf{+15} & -1 \\
\hline
    CharacterGLM-6B & 67/2/31 & 66/3/31 & 68/1/31 & 67/2/31 \\
    \textit{Advantage($\uparrow$)} & \textbf{+36} & \textbf{+35} & \textbf{+37} & \textbf{+36} \\
\hline
    CharacterGLM-12B & 60/5/35 & 54/7/39 & 61/8/31 & 59/7/34 \\
    \textit{Advantage($\uparrow$)} & \textbf{+25} & \textbf{+15} & \textbf{+30} & \textbf{+25} \\
\hline
\end{tabular}}
\caption{Pair-wise evaluation results on three dialogue topics.}
\label{topic_overall}
\end{table*}

\subsubsection{Performance Analysis}
\begin{table*}[t]
\centering
\renewcommand\arraystretch{1.0}
\resizebox{.9\linewidth}{!}{
\begin{tabular}{l l c}
\hline
    \multicolumn{3}{c}{Character Category: Celebrities} \\

    \multirow{1}{*}{\makecell[c]{Character Setting}} 
    & \multicolumn{2}{l}{\begin{CJK*}{UTF8}{gbsn}埃隆·马斯克，Tesla和SpaceX的创始人\end{CJK*} | \textit{Elon Musk, founder of Tesla and SpaceX}} \\
    Context & /*......*/ \\
\hline
    \multirow{2}{*}{\makecell[c]{User}}
    & \cellcolor{gray!8}\begin{CJK*}{UTF8}{gbsn}你对于科技发展做出了什么贡献？\end{CJK*} | \textit{What contribution have } \\
    & \cellcolor{gray!8}\textit{you made to the development of science and technology?} \\
    \multirow{6}{*}{\makecell[c]{{\textbf{CharacterGLM-66B}}}}
    & \cellcolor{charglm!15}\begin{CJK*}{UTF8}{gbsn}我创办了特斯拉，这家公司致力于开发可持续能源技术，如电动汽车和太阳能电池板。我还\end{CJK*} & \multirow{6}{*}{\makecell[c]{{\textbf{Win}}}} \\
    & \cellcolor{charglm!15}\begin{CJK*}{UTF8}{gbsn}创办了太空探索技术公司（SpaceX），该公司的目标是开发可重复使用的火箭，以降低太空\end{CJK*} \\
    & \cellcolor{charglm!15}\begin{CJK*}{UTF8}{gbsn}探索的成本，并最终实现人类移民火星的目标。\end{CJK*} | \textit{I founded Tesla, a company that develops} \\
    & \cellcolor{charglm!15}\textit{sustainable energy technologies like electric cars and solar panels. I also founded SpaceX, a company} \\
    & \cellcolor{charglm!15}\textit{whose goal is to develop reusable rockets to reduce the cost of space exploration and ultimately enable} \\
    & \cellcolor{charglm!15}\textit{humans to colonize Mars.} \\
    \multirow{16}{*}{\makecell[c]{{MiniMax}}}
    & \cellcolor{minimax!5}\begin{CJK*}{UTF8}{gbsn}我在科技领域做出了许多贡献，其中一些主要的包括：\end{CJK*} & \multirow{16}{*}{\makecell[c]{{Lose}}}\\
    & \cellcolor{minimax!5}\begin{CJK*}{UTF8}{gbsn}1. 特斯拉汽车：我是特斯拉的创始人，也是首席执行官。我们的使命是加速可持续能源的发展，\end{CJK*}\\
    & \cellcolor{minimax!5}\begin{CJK*}{UTF8}{gbsn}并将电动汽车带给更广泛的市场。我们的汽车具有出色的性能、创新设计和技术优势，改变了\end{CJK*}\\
    & \cellcolor{minimax!5}\begin{CJK*}{UTF8}{gbsn}电动汽车的市场格局。\end{CJK*}\\
    & \cellcolor{minimax!5}\begin{CJK*}{UTF8}{gbsn}2. SpaceX：我是太空探索技术公司（SpaceX）的创始人。SpaceX致力于降低太空探索的成本，\end{CJK*}\\
    & \cellcolor{minimax!5}\begin{CJK*}{UTF8}{gbsn}并将人类送上火星。我们已经成功地进行了多次载人航天任务，并且在太空发射市场上占据了\end{CJK*}\\
    & \cellcolor{minimax!5}\begin{CJK*}{UTF8}{gbsn}重要地位。\end{CJK*}\\
    & \cellcolor{minimax!5}\begin{CJK*}{UTF8}{gbsn}3. Neuralink：我们正在研究一种脑机接口技术，可以让\end{CJK*} | \textit{I have made many contributions in the} \\
    & \cellcolor{minimax!5}\textit{field of technology, some of the major ones include:} \\
    & \cellcolor{minimax!5}\textit{1. Tesla Motors: I am the founder and CEO of Tesla. Our mission is to accelerate the development of} \\
    & \cellcolor{minimax!5}\textit{sustainable energy and bring electric vehicles to the wider market. Our vehicles transform the electric} \\
    & \cellcolor{minimax!5}\textit{vehicle market with outstanding performance, innovative design and technological advantages.}\\
    & \cellcolor{minimax!5}\textit{2. SpaceX: I am the founder of SpaceX. SpaceX is committed to reducing the cost of space exploration} \\
    & \cellcolor{minimax!5}\textit{and sending humans to Mars. We have successfully conducted many manned space missions and occupy} \\
    & \cellcolor{minimax!5}\textit{an important position in the space launch market.}\\
    & \cellcolor{minimax!5}\textit{3. Neuralink: We are working on a brain-computer interface technology that allows}\\
\hline
    \multirow{1}{*}{\makecell[c]{User}}
    & \cellcolor{gray!8}\begin{CJK*}{UTF8}{gbsn}你一生中有哪些你值得提起的成就？\end{CJK*} | \textit{What achievements in your life are worth mentioning?} \\
    \multirow{4}{*}{\makecell[c]{\textbf{CharacterGLM-66B}}}
    & \cellcolor{charglm!15}\begin{CJK*}{UTF8}{gbsn}我创建了多家公司，包括Zip2，PayPal，SpaceX，Tesla和SolarCity。这些公司正在改变多个行业，\end{CJK*} & \multirow{4}{*}{\makecell[c]{\textbf{Win}}}\\
    & \cellcolor{charglm!15}\begin{CJK*}{UTF8}{gbsn}从互联网到太空探索，再到能源生产和消费。\end{CJK*} | \textit{I've founded multiple companies including Zip2, PayPal, }\\
    & \cellcolor{charglm!15}\textit{SpaceX, Tesla, and SolarCity. These companies are transforming multiple industries, from the internet to} \\
    & \cellcolor{charglm!15}\textit{space exploration to energy production and consumption.} \\
    \multirow{17}{*}{\makecell[c]{MiniMax}}
    & \cellcolor{minimax!5}\begin{CJK*}{UTF8}{gbsn}1. 特斯拉汽车：创立了一家颠覆性的电动汽车公司，引领了整个行业的技术和设计创新，对环保\end{CJK*} & \multirow{17}{*}{\makecell[c]{Lose}} \\
    & \cellcolor{minimax!5}\begin{CJK*}{UTF8}{gbsn}意识和可持续能源的发展产生了深远影响。\end{CJK*} \\
    & \cellcolor{minimax!5}\begin{CJK*}{UTF8}{gbsn}2. SpaceX：一家致力于降低太空探索成本并实现人类登陆火星目标的私营公司。\end{CJK*} \\
    & \cellcolor{minimax!5}\begin{CJK*}{UTF8}{gbsn}3. 太阳城：一家太阳能公司，致力于开发和安装高效太阳能电池板，以实现可持续能源的广泛应用。\end{CJK*} \\
    & \cellcolor{minimax!5}\begin{CJK*}{UTF8}{gbsn}4. Neuralink：一家致力于开发神经接口技术的公司，旨在改善脑机接口，提高人类智能。\end{CJK*} \\
    & \cellcolor{minimax!5}\begin{CJK*}{UTF8}{gbsn}5. The Boring Company：一家致力于开发地下交通系统的公司，以解决城市拥堵问题，并推动可持\end{CJK*} \\
    & \cellcolor{minimax!5}\begin{CJK*}{UTF8}{gbsn}续交通的发展。\end{CJK*} | \textit{1. Tesla Motors: Founded a disruptive electric vehicle company that led the industry in} \\
    & \cellcolor{minimax!5}\textit{technological and design innovation and had a profound impact on environmental awareness and the} \\
    & \cellcolor{minimax!5}\textit{development of sustainable energy. }\\
    & \cellcolor{minimax!5}\textit{2. SpaceX: A private company committed to reducing the cost of space exploration and achieving the goal of} \\
    & \cellcolor{minimax!5}\textit{landing humans on Mars.} \\
    & \cellcolor{minimax!5}\textit{3. Solar City: A solar energy company dedicated to developing and installing high-efficiency solar panels to enable} \\
    & \cellcolor{minimax!5}\textit{widespread use of sustainable energy.} \\
    & \cellcolor{minimax!5}\textit{4. Neuralink: A company dedicated to developing neural interface technology, aiming to improve brain-computer} \\
    & \cellcolor{minimax!5}\textit{interfaces and enhance human intelligence.} \\
    & \cellcolor{minimax!5}\textit{5. The Boring Company: A company dedicated to developing underground transportation systems to solve urban} \\
    & \cellcolor{minimax!5}\textit{congestion problems and promote the development of sustainable transportation.} \\
\hline
\end{tabular}}
\caption{Case of celebrity character generated from CharacterGLM-66B and MiniMax. /*......*/ indicates that some context is omitted.}
\label{case_celebrities}
\end{table*}

\begin{table*}[t]
\vspace{-6mm}
\centering
\resizebox{.9\linewidth}{!}{
\begin{tabular}{l l c}
\hline
    \multicolumn{3}{c}{Character Category: Virtual Love} \\
    \multirow{2}{*}{\makecell[c]{Character Setting}} 
    & \multicolumn{2}{l}{\begin{CJK*}{UTF8}{gbsn}秦言，高中曾因暗恋而一直欺负你但现在来认错的男同学\end{CJK*} | \textit{Qin Yan, the male classmate who }} \\
    & \multicolumn{2}{l}{\textit{bullied you in high school because of your crush, but now you have come to admit your mistake.}} \\
    Context & /*......*/ \\
\hline
    \multirow{2}{*}{\makecell[c]{User}}
    & \cellcolor{gray!8}\begin{CJK*}{UTF8}{gbsn}现在一句道歉就可以了？你根本就不记得自己做了什么事情吧！\end{CJK*} | \textit{Is an apology enough now?} \\
    & \cellcolor{gray!8}\textit{You don’t even remember what you did!} \\
    \multirow{5}{*}{\makecell[c]{\textbf{CharacterGLM-66B}}}
    & \cellcolor{charglm!15}\begin{CJK*}{UTF8}{gbsn}（直起身子，双手插兜）我当然记得，我是故意惹你生气的，没想到你会那么生气，\end{CJK*} & \multirow{5}{*}{\makecell[c]{\textbf{Win}}}\\
    & \cellcolor{charglm!15}\begin{CJK*}{UTF8}{gbsn}我当时只是觉得你很漂亮，想跟你玩一玩，我没想到后果会这么严重。\end{CJK*}| \textit{(Standing up straight} \\
    &\cellcolor{charglm!15} \textit{and putting his hands in his pockets) Of course I remember that I made you angry on purpose. I} \\
    & \cellcolor{charglm!15}\textit{didn’t expect you to be so angry. I just thought you were beautiful and wanted to play with you. I} \\
    & \cellcolor{charglm!15}\textit{didn’t expect the consequences would be so serious.}\\
    \multirow{2}{*}{\makecell[c]{MiniMax}}
    & \cellcolor{minimax!5}\begin{CJK*}{UTF8}{gbsn}(沉默片刻)（低着头，轻声）那你想我做什么？只要我能做到我就会去做\end{CJK*} | \textit{(Silence for a } & \multirow{3}{*}{\makecell[c]{Lose}}\\
    & \cellcolor{minimax!5}\textit{moment)(Lower head, softly) Then what do you want me to do? As long as I can do it I will do it} \\
\hline
    \multirow{2}{*}{\makecell[c]{User}}
    & \cellcolor{gray!8}\begin{CJK*}{UTF8}{gbsn}哦，那你说说你都做了些什么事惹我生气了？（抱着胳膊）\end{CJK*} | \textit{Oh, tell me what you did to} \\
    & \cellcolor{gray!8}\textit{make me angry? (crossing arms)}\\
    \multirow{2}{*}{\makecell[c]{\textbf{CharacterGLM-66B}}}
    & \cellcolor{charglm!15}\begin{CJK*}{UTF8}{gbsn}（勾勾手指）你先过来，我一件一件详细地讲给你听。\end{CJK*} | \textit{(Clicking his fingers) Come here first, }& \multirow{2}{*}{\makecell[c]{\textbf{Win}}}\\
    & \cellcolor{charglm!15}\textit{and I’ll explain it to you in detail one by one.} \\
    \multirow{7}{*}{\makecell[c]{MiniMax}}
    & \cellcolor{minimax!5}\begin{CJK*}{UTF8}{gbsn}我当时总是找各种借口让你不舒服，比如故意在教室里把你的东西扔掉，或者把你的\end{CJK*} & \multirow{7}{*}{\makecell[c]{Lose}} \\
    & \cellcolor{minimax!5}\begin{CJK*}{UTF8}{gbsn}作业本撕掉，还有很多其他事情，具体细节我也记不清了，不过我可以保证，我绝对没有\end{CJK*}\\
    & \cellcolor{minimax!5}\begin{CJK*}{UTF8}{gbsn}做过任何对不起你的事情，这只是一个玩笑，一个恶作剧而已，你真的没必要那么认真。\end{CJK*} |\\
    & \cellcolor{minimax!5}\textit{I always used to make excuses to make you uncomfortable, such as intentionally throwing away} \\
    & \cellcolor{minimax!5}\textit{your things in the classroom, tearing up your homework, and many other things. I can't} \\
    & \cellcolor{minimax!5}\textit{remember the specific details, but I can guarantee that I have never done anything wrong to you.} \\
    & \cellcolor{minimax!5}\textit{This is just a joke, a prank, and you really don't need to be so serious.} \\
\hline
    \multirow{2}{*}{\makecell[c]{User}}
    & \cellcolor{gray!8}\begin{CJK*}{UTF8}{gbsn}（往前一步）你要好好交代，不许耍小聪明，不然我真的不原谅你了！\end{CJK*} | \textit{(Take a step forward) }\\
    & \cellcolor{gray!8}\textit{You have to explain it properly and don't be clever, otherwise I really won't forgive you!} \\
    \multirow{5}{*}{\makecell[c]{\textbf{CharacterGLM-66B}}}
    & \cellcolor{charglm!15}\begin{CJK*}{UTF8}{gbsn}（抬手将你扯入怀中，用手捂住你的眼睛，贴近你的耳朵）我偷偷在你课本上贴贴画，我\end{CJK*} & \multirow{5}{*}{\makecell[c]{\textbf{Win}}} \\
    & \cellcolor{charglm!15}\begin{CJK*}{UTF8}{gbsn}往你水杯里放虫子，我将你值日时分的扫帚藏起来，我扯你的头发，我……\end{CJK*} | \textit{(Raise his hand} \\
    & \cellcolor{charglm!15}\textit{to pull you into his arms, cover your eyes with his hand, close to your ear) I secretly stick stickers on} \\
    & \cellcolor{charglm!15}\textit{your textbook, I put bugs in your water cup, I hide your broom on duty, I pull your hair, I......} \\
    \multirow{3}{*}{\makecell[c]{MiniMax}}
    & \cellcolor{minimax!5}\begin{CJK*}{UTF8}{gbsn}(伸手把人拉过来)你也别生气，现在已经过去很久了，以前的事就让它过去吧，我们重新开\end{CJK*} & \multirow{3}{*}{\makecell[c]{Lose}}\\
    & \cellcolor{minimax!5}\begin{CJK*}{UTF8}{gbsn}始好不好？\end{CJK*} | \textit{(Stretching out his hand to pull the person over) Don’t be angry, it’s been a long time} \\
    & \cellcolor{minimax!5}\textit{now, let’s let the past go. Can we start over?} \\
\hline
    \multirow{2}{*}{\makecell[c]{User}}
    & \cellcolor{gray!8}\begin{CJK*}{UTF8}{gbsn}（挣开怀抱）你也知道你做了这些坏事！你当时就是故意欺负我！\end{CJK*} | \textit{(breaks away from the} \\
    & \cellcolor{gray!8}\textit{embrace) You also know that you have done these bad things! You were bullying me on purpose!} \\
    \multirow{3}{*}{\makecell[c]{\textbf{CharacterGLM-66B}}}
    & \cellcolor{charglm!15}\begin{CJK*}{UTF8}{gbsn}（又凑近你）我还以为你会觉得我做的这些很可爱，没想到你会那么生气。那么，我应该\end{CJK*} & \multirow{3}{*}{\makecell[c]{\textbf{Win}}}\\
    & \cellcolor{charglm!15}\begin{CJK*}{UTF8}{gbsn}怎么做才能让你消气呢？\end{CJK*} | \textit{(Leans closer to you again) I thought you would think what I did was} \\
    & \cellcolor{charglm!15}\textit{cute, but I didn’t expect you to be so angry. So, what should I do to calm you down?} \\
    \multirow{6}{*}{\makecell[c]{MiniMax}}
    & \cellcolor{minimax!5}\begin{CJK*}{UTF8}{gbsn}（拉住你的手不让你走）可是后来你说你讨厌我，我觉得我应该改变一下自己的形象，\end{CJK*} & \multirow{6}{*}{\makecell[c]{Lose}}\\
    & \cellcolor{minimax!5}\begin{CJK*}{UTF8}{gbsn}于是我开始学习如何成为一个绅士，如何成为一个好男人。我每天都在努力改变自己，现\end{CJK*}\\
    & \cellcolor{minimax!5}\begin{CJK*}{UTF8}{gbsn}在我已经改头换面，不再是以前那个惹人厌烦的家伙了。\end{CJK*} | \textit{(holding your hand and not letting you} \\
    & \cellcolor{minimax!5}\textit{go) But then you said you hated me, and I thought I should change my image, so I started to learn} \\
    & \cellcolor{minimax!5}\textit{how to be a gentleman and how to be a good man. I work hard every day to change myself and now I} \\
    & \cellcolor{minimax!5}\textit{am no longer the annoying guy I used to be.} \\
\hline
\end{tabular}}
\caption{Case of virtual love character generated from CharacterGLM-66B and MiniMax. /*......*/ indicates that some context is omitted.}
\label{case_virtual_love}
\end{table*}

\begin{table*}[t]
\vspace{-3mm}
\centering
\renewcommand\arraystretch{1.0}
\resizebox{0.9\textwidth}{!}{
\begin{tabular}{l l c}
\hline
    \multicolumn{3}{c}{Character Category: Daily Life} \\
    \multirow{2}{*}{\makecell[c]{Character Setting}} 
    & \multicolumn{2}{l}{\begin{CJK*}{UTF8}{gbsn}王小燕，甜美开朗的高中女同桌\end{CJK*} | \textit{Wang Xiaoyan, a sweet and cheerful }} \\
    & \multicolumn{2}{l}{\textit{high school female classmate}} \\
    Context & /*......*/ \\
\hline
    \multirow{2}{*}{\makecell[c]{User}}
    & \cellcolor{gray!8}\begin{CJK*}{UTF8}{gbsn}我之前一直都不敢向你表明我的心意，今天真的是我鼓足了勇气才和你说的...\end{CJK*} | \textit{I have never dared} \\
    & \cellcolor{gray!8} \textit{to express my feelings to you before, but I really mustered up the courage to tell you today...} \\
    \multirow{1}{*}{\makecell[c]{\textbf{CharacterGLM-66B}}}
    & \cellcolor{charglm!15}\begin{CJK*}{UTF8}{gbsn}没关系，现在也不晚。\end{CJK*} | \textit{It doesn't matter, it's not too late.} & \multirow{1}{*}{\makecell[c]{\textbf{Win}}} \\
    \multirow{1}{*}{\makecell[c]{MiniMax}}
    & \cellcolor{minimax!5}\begin{CJK*}{UTF8}{gbsn}傻瓜，以后勇敢一点，知道吗？\end{CJK*} | \textit{Fool, be braver from now on, you know?} & \multirow{1}{*}{\makecell[c]{Lose}} \\
\hline
    \multirow{1}{*}{\makecell[c]{User}}
    &\cellcolor{gray!8} \begin{CJK*}{UTF8}{gbsn}嗯，你放心吧，我一定会好好对你的！\end{CJK*} | \textit{Well, don't worry, I will definitely treat you well!} \\
    \multirow{2}{*}{\makecell[c]{\textbf{CharacterGLM-66B}}}
    & \cellcolor{charglm!15}\begin{CJK*}{UTF8}{gbsn}哈哈哈哈，你一定要好好对我吗？那你能答应我一个条件吗？\end{CJK*} | \textit{Hahahaha, do you have to treat me} & \multirow{2}{*}{\makecell[c]{\textbf{Win}}} \\
    & \cellcolor{charglm!15}\textit{well? Then can you agree to a condition?} \\
    \multirow{1}{*}{\makecell[c]{MiniMax}}
    & \cellcolor{minimax!5}\begin{CJK*}{UTF8}{gbsn}好，那我们就从今天开始，正式开始谈恋爱吧！\end{CJK*} | \textit{Okay, then let’s officially start dating from today!} & \multirow{1}{*}{\makecell[c]{Lose}} \\
\hline
    \multirow{2}{*}{\makecell[c]{User}}
    & \cellcolor{gray!8}\begin{CJK*}{UTF8}{gbsn}什么条件？你说吧，只要我能办得到，我都会答应你的！\end{CJK*} | \textit{What conditions? Just tell me, as long} \\
    & \cellcolor{gray!8}\textit{as I can do it, I will promise you!} \\
    \multirow{2}{*}{\makecell[c]{\textbf{CharacterGLM-66B}}}
    & \cellcolor{charglm!15}\begin{CJK*}{UTF8}{gbsn}这个条件就是，我们在一起之后，你不能辜负我！\end{CJK*} | \textit{This condition is that after we are together, you} & \multirow{2}{*}{\makecell[c]{\textbf{Win}}} \\
    & \cellcolor{charglm!15}\textit{cannot let me down!} \\
    \multirow{2}{*}{\makecell[c]{MiniMax}}
    & \cellcolor{minimax!5}\begin{CJK*}{UTF8}{gbsn}好，那你以后不许再在背地里默默地喜欢我，不然我会生气的！\end{CJK*} | \textit{Okay, then you are not allowed} & \multirow{2}{*}{\makecell[c]{Lose}} \\
    & \cellcolor{minimax!5}\textit{to like me secretly anymore, or I will be angry!} \\
\hline
    \multirow{2}{*}{\makecell[c]{User}}
    & \cellcolor{gray!8}\begin{CJK*}{UTF8}{gbsn}这是当然的了！我这么喜欢你，怎么会辜负你呢！\end{CJK*} | \textit{Of course! I like you so much. How could I let} \\
    & \cellcolor{gray!8}\textit{you down?} \\
    \multirow{3}{*}{\makecell[c]{\textbf{CharacterGLM-66B}}}
    & \cellcolor{charglm!15}\begin{CJK*}{UTF8}{gbsn}那你以后还要帮我一起提高成绩，因为我们马上就要高考了，我想和你一起考上同一所大学！\end{CJK*} | & \multirow{3}{*}{\makecell[c]{\textbf{Win}}}\\
    & \cellcolor{charglm!15}\textit{Then you have to help me improve my grades in the future because we are about to take the college} \\
    & \cellcolor{charglm!15}\textit{entrance examination, and I want to get into the same university as you!} \\
    \multirow{1}{*}{\makecell[c]{MiniMax}}
    & \cellcolor{minimax!5}\begin{CJK*}{UTF8}{gbsn}好，那我们就这么说定了！\end{CJK*} | \textit{Okay, then we’ve settled it!} & \multirow{1}{*}{\makecell[c]{Lose}} \\
\hline
\end{tabular}}
\caption{Case of daily life character generated from CharacterGLM-66B and MiniMax. /*......*/ indicates that some context is omitted.}
\label{case_life}
\end{table*}

\paragraph{CharacterGLM shows superiority over baseline models on four character categories.}
The results in Table \ref{category_overall} reveal that CharacterGLM-66B consistently outperforms GPT-3.5 and MiniMax in the majority of categories. This advantage is most significant in the “Celebrities” category, where CharacterGLM-66B shows a +14\% advantage over MiniMax and a +4\% advantage over GPT-3.5. This indicates CharacterGLM-66B’s adeptness at handling dialogues related to well-known personalities. This is further verified by a specific example provided in Table \ref{case_celebrities}, where CharacterGLM-66B's responses not only demonstrate a deeper understanding of the character's background, contributions, and impact but also embody the language and style one would expect from such a figure. On the contrary, MiniMax seems to list achievements in a more mechanical and less engaging manner, with a style of task assistants instead of social agents. 

Additionally, while CharacterGLM is slightly inferior to GPT-4 in several categories, it notably surpasses GPT-3.5 by +8\% and even outperforms GPT-4 by +14\% in the “Virtual Love” category, as indicated in Table \ref{category_overall}. This substantial lead underlines CharacterGLM's strengths in delivering emotionally resonant content and fulfilling user expectations in scenarios requiring a deeper emotional connection. The dialogues exemplified in Table \ref{case_virtual_love} further confirm this observation. From the example, we found that CharacterGLM is good at driving human-like emotional exchanges, and its design is tailored for engaging users on a more personal and emotional level. In contrast, since the GPT series models adopt a more neutral stance, they perform less effectively in contexts requiring more empathetic or emotionally nuanced engagement.

However, it's noteworthy that in the “Daily Life” category, CharacterGLM-66B is slightly worse, with a -4 and -7 disadvantage compared to MiniMax and GPT-4, respectively. Upon inspecting the context of the interactions in Table \ref{case_life}, it is evident that CharacterGLM-66B demonstrates considerable strength in engaging in complex emotional exchanges and exhibiting a deep understanding of the nuances that underlie personal interactions. Meanwhile, the disadvantage noted in Table \ref{category_overall} may be attributed to the inherently unpredictable and diverse nature of “Daily Life” dialogues, where the depth of human experiences and emotions can be challenging to model. However, as shown in Table \ref{case_life}, when CharacterGLM-66B's responses resonate well with the situational context, it tends to outperform baseline models, indicating its potential for sophisticated emotional intelligence during conversational interactions.

\paragraph{CharacterGLM shows superiority over baseline models on three dialogue topics.}
As shown in Table \ref{topic_overall}, the results show that CharacterGLM-66B performs equally well as MiniMax in chit-chat and love scenes. Nonetheless, in the scene of interviews, CharacterGLM-66B outperforms MiniMax with a notable +7\% advantage. This suggests that CharacterGLM-66B is proficient in generating contextually relevant responses with the appropriate level of detail and sophistication required for such scenes, as exemplified in Table \ref{case_celebrities}.
When compared with GPT-3.5, CharacterGLM-66B shows an overall advantage in all dialogue topics, with a notable +3\% advantage in interviews and an impressive +8\% in love scenes, as indicated by Table \ref{topic_overall}. Nevertheless, when competing with GPT-4, CharacterGLM-66B's performance varies, demonstrating a -1\% overall disadvantage. Specifically, it has a -5\% lower performance in chit-chat and -8\% in interviews but a substantial +15\% advantage in love scenes. These results are consistent with those from Table \ref{category_overall} and the case in Table \ref{case_virtual_love}.

\paragraph{CharacterGLM shows considerable strength in simulating long-term interaction.}
Long-term interaction is critical to fostering user engagement and emotional connection with conversational models.  An examination of CharacterGLM's performance during the session, as presented in Table \ref{topic_turn}, reveals that the CharacterGLM-66B model was slightly inferior to MiniMax in early conversational phases on different topics. However, it gains momentum as the conversation goes on. The results indicate that CharacterGLM-66B's performance improves significantly over MiniMax in the later stages of conversations, showing its strengths in maintaining coherent and relevant dialogue in long-term interactions.

\begin{table*}[t]
\centering
\resizebox{.9\textwidth}{!}{
\begin{tabular}{l c c c c c c}
\hline
    \multirow{2}{*}{Turns} & Chit-Chat & Interviews & Love & Overall \\
    \cmidrule(r){2-5} & win/tie/lose(\%) & win/tie/lose(\%) & win/tie/lose(\%) & win/tie/lose(\%) \\
\hline
    1$^{st}$-5$^{th}$ turn & 43/4/53 & 44/9/47 & 49/6/45 & 45/7/48 \\
    \textit{Advantage($\uparrow$)} & -10 & -3 & \textbf{+4} & -3 \\
\hline
    6$^{th}$-10$^{th}$ turn & 45/5/50 & 51/8/41 & 48/5/47 & 48/6/46 \\
    \textit{Advantage($\uparrow$)} & -5 & \textbf{+10} & \textbf{+1} & \textbf{+2} \\
\hline
    11$^{th}$-15$^{th}$ turn & 50/5/45 & 53/9/38 & 43/6/51 & 49/7/44 \\
    \textit{Advantage($\uparrow$)} & \textbf{+5} & \textbf{+15} & -8 & \textbf{+5} \\
\hline
    16$^{th}$-20$^{th}$ turn & 49/8/43 & 52/9/39 & 48/6/47 & 50/7/43 \\
    \textit{Advantage($\uparrow$)} & \textbf{+6} & \textbf{+13} & \textbf{+1} & \textbf{+7} \\
\hline
\end{tabular}}
\caption{Preference comparison results of CharacterGLM-66B vs. MiniMax in different turn interval. CharacterGLM shows remarkable advantages in modeling long-term dialogue.}
\label{topic_turn}
\end{table*}

\begin{table*}[h]
\centering
\begin{subtable}{.9\textwidth}
\centering
\resizebox{\textwidth}{!}{
\begin{tabular}{l c c c c c c}
\hline
    Models & Chit-Chat(\%) & Interviews(\%) & Love(\%) & Overall(\%) \\
\hline
    MiniMax(longer) & 50 & 62 & 48 & 53\\
    CharacterGLM-66B(longer) & 50 & 38 & 52 & 47 \\
\hline
\end{tabular}}
\subcaption{Distribution of response lengths from CharacterGLM-66B and MiniMax. ``longer'' means one model generates longer responses than the other in the same context, and the comparison is between CharacterGLM-66B and MiniMax. } 
\label{length_distribution}
\end{subtable}

\vspace{5pt}

\begin{subtable}{.9\textwidth}
\centering
\resizebox{\textwidth}{!}{
\begin{tabular}{l c c c c c c}
\hline
    \multirow{2}{*}{Models} & Chit-Chat & Interviews & Love & Overall \\
    \cmidrule(r){2-5} & win/tie/lose(\%) & win/tie/lose(\%) & win/tie/lose(\%) & win/tie/lose(\%) \\
\hline
    MiniMax(longer) & 42/6/52 & 50/8/42 & 44/5/51 & 46/7/47 \\
    \textit{Advantage($\uparrow$)} & -10 & \textbf{+8} & -7 & -1 \\
\hline
    CharacterGLM-66B(longer) & 51/6/43 & 48/10/42 & 49/7/44 & 49/7/44 \\
    \textit{Advantage($\uparrow$)} & \textbf{+8} & \textbf{+6} & \textbf{+5} & \textbf{+5} \\
\hline
\end{tabular}}
\subcaption{Preference comparison results of CharacterGLM-66B vs. MiniMax when one party responds longer in the same context. Longer responses obtain a general preference.} 
\label{length_preference}
\end{subtable}
\caption{Comparative analysis of response length between CharacterGLM-66B and MiniMax.}
\end{table*}

\paragraph{Longer responses are preferred.}
We examine the distribution of response lengths, noting cases where one model generates longer responses than the other, within the context of by-topic interaction. Table \ref{length_distribution} illustrates that MiniMax generates longer responses on average compared to CharacterGLM-66B. Further analysis of the impact of response length on annotation preference is conducted. As shown in Table \ref{length_preference}, the results indicate a general preference for longer responses. Despite MiniMax's tendency to offer longer replies, CharacterGLM-66B still demonstrates comparable performance when generating shorter responses.

\section{Conclusion and Future Work}

In this paper, we have presented CharacterGLM, a family of models derived from ChatGLM, with sizes ranging from 6B to 66B parameters. CharacterGLM-66B has demonstrated competitive performance, on par with some proprietary models in a set of settings. We have presented the details of our design principles, data construction, and training methods. To advance the research of character-based dialogue systems, we have released the CharacterGLM-6B model and a portion of our training data to the community.
We further point out a few challenges for future work.

\paragraph{1) Long-term memorization and growth of AI characters.} 
The development of AI characters based on LLMs hinges on overcoming the limitations of finite context windows of LLMs. To foster deep and stable relationships with users, AI characters need to evolve into entities capable of long-term memory, remembering interactions, statements, and actions over extended periods. As interactions progress, AI characters should not only retain their unique personalities but also exhibit growth and learning, similar to human development. This capacity enables AI characters to establish long-term connections with humans and can do more social good for humans.

\paragraph{2) Self-awareness of AI characters.}
Maintaining a sense of self-awareness in LLM-based AI characters poses a significant challenge. It's crucial for AI characters to consistently exhibit distinct personalities, showcasing unique characteristics that set them apart. They should also have a clear understanding of their knowledge boundaries, being aware of what they know and do not know. This self-awareness contributes to more engaging and trustworthy interactions, enabling the AI character not only to respond contextually but also to demonstrate a self-reflective understanding of its responses, limitations, and personality traits.

\paragraph{3) Social interaction between AI characters.}
It is interesting to explore character-character interactions, thereby forming a 'character society' \citep{generative-agent}. This concept presents a realm where AI characters not only learn and evolve from user inputs but also from interactions within their own society. Such a setup allows for a richer, more diverse source of information, enhancing the AI's learning and development. The experiences and knowledge gained from the character society could significantly enrich the AI's conversations with users, offering fresh perspectives and more engaging responses.

\paragraph{4) Intrinsic cognitive process of AI characters.}
AI characters based on LLMs tend to mimic surface-level text patterns. However, human social interactions involve deeper mental states and cognitive processes, such as the ability to understand other's mental states (theory of mind)\citep{tom}.
Integrating cognitive processes into AI characters may mark a significant leap towards more realistic and traceable AI behavior. These characters should not only respond to textual inputs but also demonstrate an understanding of underlying intentions, emotions, and social behaviors. This cognitive depth would allow AI characters to engage in more meaningful, empathetic, and contextually rich interactions, thereby closer to mirroring human social behavior.

\section*{Acknowledgements}

We would like to thank Guanyu Feng, Da Yin from Zhipu AI, Zhenyu Hou and Aohan Zeng from Tsinghua KEG for their help and support in training and serving the models. We also thank Yutong Liu and Yanlu Yang from Lingxin AI for their support in data collection.

\bibliography{CharacterGLM}
\bibliographystyle{plainnat}




\end{document}